\documentclass[acmtog,nonacm]{acmart}
\AtBeginDocument{%
  \providecommand\BibTeX{{%
    \normalfont B\kern-0.5em{\scshape i\kern-0.25em b}\kern-0.8em\TeX}}}

\usepackage{hyperref}
\usepackage{multirow}
\usepackage{graphicx}
\usepackage{bbding}
\usepackage{wrapfig}
\usepackage{subcaption}

\AtBeginDocument{%
  \providecommand\BibTeX{{%
    \normalfont B\kern-0.5em{\scshape i\kern-0.25em b}\kern-0.8em\TeX}}}

\acmISBN{978-1-4503-XXXX-X/18/06}

\begin{document}

\title{Den-SOFT: Dense \textbf{S}pace-\textbf{O}riented Light \textbf{F}ield Datase\textbf{T} for 6-DOF Immersive Experience}

\author{Xiaohang Yu}
\authornote{Both authors contributed equally to this research.}
\author{Zhengxian Yang}
\authornotemark[1]
\author{Shi Pan}
\authornotemark[1]
\affiliation{%
  \institution{Tsinghua University}
  \city{Beijing}
  \country{China}
}
\email{xiaoyu912huang@163.com}
\email{zx-yang23@mails.tsinghua.edu.cn}
\email{yixingpanshi@outlook.com}

\author{Yuqi Han}
\affiliation{%
  \institution{Tsinghua University}
  \city{Beijing}
  \country{China}
}

\author{Haoxiang Wang}
\affiliation{%
  \institution{Tsinghua University}
  \city{Beijing}
  \country{China}
}
\email{whx22@mails.tsinghua.edu.cn}

\email{hyq_2011061510@163.com}

\author{Jun Zhang}
\affiliation{%
  \institution{Tsinghua University}
  \city{Beijing}
  \country{China}
}
\email{}
\author{Shi Yan}
\affiliation{%
  \institution{Tsinghua University}
  \city{Beijing}
  \country{China}
}
\email{}
\author{Borong Lin}
\affiliation{%
  \institution{Tsinghua University}
  \city{Beijing}
  \country{China}
}
\email{linbr@tsinghua.edu.cn}

\author{Lei Yang}
\affiliation{%
  \institution{SenseTime}
  \city{Beijing}
  \country{China}
}
\email{yanglei@sensetime.com}

\author{Tao Yu}
\authornote{Corresponding Author.}
\author{Lu Fang}
\authornotemark[2]
\affiliation{%
  \institution{Tsinghua University}
  \city{Beijing}
  \country{China}
}
\email{ytrock@tsinghua.edu.cn}
\email{fanglu@tsinghua.edu.cn}


\renewcommand{\shortauthors}{Xiaohang Yu and Zhengxian Yang, et al.}

\begin{abstract}
  We have built a custom mobile multi-camera large-space dense light field capture system, which provides a series of high-quality and sufficiently dense light field images for various scenarios. 
  Our aim is to contribute to the development of popular 3D scene reconstruction algorithms such as IBRnet, NeRF, and 3D Gaussian splitting. 
  More importantly, the collected dataset, which is much denser than existing datasets, may also inspire space-oriented light field reconstruction, which is potentially different from object-centric 3D reconstruction, for immersive VR/AR experiences. 
  We utilized a total of 40 GoPro 10 cameras, capturing images of 5k resolution. The number of photos captured for each scene is no less than 1000, and the average density (view number within a unit sphere) is 134.68. 
  It is also worth noting that our system is capable of efficiently capturing large outdoor scenes. Addressing the current lack of large-space and dense light field datasets, we made efforts to include elements such as sky, reflections, lights and shadows that are of interest to researchers in the field of 3D reconstruction during the data capture process. Finally, we validated the effectiveness of our provided dataset on three popular algorithms and also integrated the reconstructed 3DGS results into the Unity engine, demonstrating the potential of utilizing our datasets to enhance the realism of virtual reality (VR) and create feasible interactive spaces. The dataset is available at: \href{https://metaverse-ai-lab-thu.github.io/Den-SOFT/}{https://metaverse-ai-lab-thu.github.io/Den-SOFT/}
\end{abstract}


\begin{teaserfigure}
  \includegraphics[width=\textwidth]{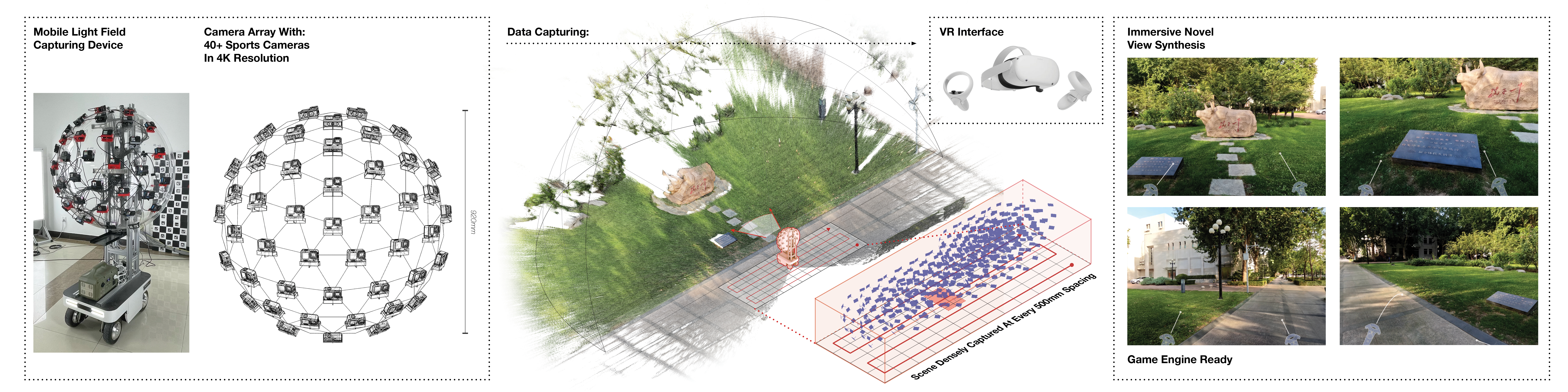}
  \caption{Den-SOFT makes real-time 6DoF immersive experience in VR possible. We have utilized a self-designed multi-camera device named "Compound Eye" to capture a dataset of images and videos from multiple scenes, with a particular focus on large-scale outdoor environments. The quality and viewpoint density of this dataset may represent the current state-of-the-art within the public domain as we validated the effectiveness on popular algorithms and integrated it with VR engine.}
  \Description{Our process from capturing to VR.}
  \label{fig:teaser}
\end{teaserfigure}

\maketitle

\section{Introduction}
With the widespread application of deep neural networks and the rapid advancement of virtual reality (VR) technology, digital reconstruction of the real world has garnered unprecedented attention. This interest extends beyond small-scale personal item design and human reconstruction to the virtual exploration of interior decorations and expansive outdoor scenes. Implicit neural radiance field rendering methods, epitomized by NeRF~\cite{mildenhall2021nerf}, have brought research in this field to a climax. However, despite theoretical advancements in 3D reconstruction, the field still faces significant challenges due to the high complexity of the algorithms and the scarcity of high-quality datasets.
This is especially true in the area of large-space light field reconstruction closely related to VR technology, where there is an urgent need for datasets that covers a sufficiently large range of scenes, with sufficient quality and richness to support algorithm research and testing for truly immersive six degrees of freedom (6DoF) VR experiences.

In recent years, although scholars have invested considerable effort in building high-quality 3D scene datasets, the significant improvements in VR device performance have revealed deficiencies in existing datasets in multiple aspects. These limitations have, to some extent, impacted the datasets' ability to meet the growing demands for VR experiences. Specifically:

(1) Existing datasets cannot meet the needs for high-precision and high-fidelity reconstruction of large-scale static scenes. Many public datasets have resolutions below 2K and low capture densities, which are not conducive to high-precision scene detail reconstruction. Moreover, most datasets, such as Tanks $\&$ Temples~\cite{knapitsch2017tanks} and NeRF~\cite{mildenhall2021nerf}, are essentially object-centric with very small coverage of the background details, which are essential for immersive 6DoF VR experiences.

(2) Existing datasets do not support the realization of 6DoF immersive experiences in VR. Works represented by Immersive Light Fields~\cite{broxtonImmersiveLightField2020b} and Zip-NeRF~\cite{barron2023zip} either have a fixed-point capture or are limited by the capture path, resulting in a very limited explorable space range (1 cubic meter) in VR after reconstruction.

To address these issues, our dataset \textbf{Den-SOFT} features:
\begin{itemize}
\item {\textbf{High resolution}}: Each viewpoint has a resolution of up to 5K.
\item {\textbf{High-density capture}}: Using the number of viewpoints within a unit sphere as the density metric, the average density within the capture volume of our Den-SOFT dataset reaches 134, which significantly surpasses existing datasets. 
\item {\textbf{Extensive coverage of indoor and outdoor scenes}}: Particularly for outdoor scenes, the dataset exhibits rich foreground and background details.
\end{itemize}

We have also evaluated the effectiveness of our dataset and studied the performance of current popular algorithms on it. Our main contributions are summarized as follows:

(1) In response to the problems with existing 3D scene reconstruction datasets, we designed a mobile capture system equipped with 46 high-resolution synchronized cameras mounted on a remote-controlled car. This system is capable of efficiently and densely capturing large static scenes in a short time and dynamic large scenes over an extended period.

(2) We present a new dataset that features high resolution, extremely high capture density, and covers a broad range of indoor and outdoor scenes, especially rich outdoor scenes, to support the realization of 6DoF VR experiences.

(3) Utilizing this dataset, we tested three existing paradigms of light field reconstruction methods and conducted a preliminary analysis of their strengths and weaknesses, which is expected to further drive the improvement and development of these algorithms. By using Unity as a bridge connected to the headset, we completed the entire process from data capture to scene application, demonstrating the potential of the dataset in practical VR applications. 
The highest capturing density also clarifies its potential role in promoting or evaluating research on other 3D scene reconstruction algorithms such as dynamic scene reconstruction, semantic segmentation, and understanding.

\section{Related Work}

The rapid development of capture devices and simulation techniques has led to the proposal of several benchmarks in the field of 3D reconstruction in recent years. In this section, we'll delve into an organized overview of the existing common datasets in the field of 3D reconstruction, categorizing them into two main types: Monocular-Camera-Based Dataset and Multi-Camera-Based Dataset.

\subsection{Monocular-Camera-Based Dataset} Due to their ease of setup, straightforward operation, and cost-effectiveness, monocular camera systems are predominantly used in the data acquisition of most current object-centric or indoor/outdoor scene datasets~\cite{jensen2014large,dai2017scannet,knapitsch2017tanks,bertel2020omniphotos,yoon2020novel,mildenhall2021nerf,philip2021free,barron2022mip,barron2023zip}. As we continue to strive for the creation of more realistic and interactive virtual worlds, the need for high-resolution, high-density, and large-scale scene datasets has become increasingly paramount. This demand is growing in both academic and industrial sectors, particularly in the realms of Virtual Reality (VR) and Augmented Reality (AR) applications.

Datasets such as Tanks $\&$ Temples~\cite{knapitsch2017tanks}, OmniPhotos~\cite{bertel2020omniphotos}, NeRF~\cite{mildenhall2021nerf}, and Mip-NeRF~\cite{barron2022mip}, which employ monocular cameras to capture a multitude of multi-view images and video sequences of objects or scenes, have made significant contributions to the research and development of 3D object and scene reconstruction. However, their outside-looking-in collection methodology limits them to capturing static scenes from a fixed point. This means the range of 3D scenes that can be reconstructed by algorithms is restricted, hindering their ability to support novel view synthesis tasks on a larger scale. Additionally, these types of datasets often prioritize the collection of foreground objects, frequently neglecting aspects of the background and distant views. This oversight limits their ability to provide a comprehensive sense of immersion for VR applications.

On the other hand, inside-looking-out acquisition methods such as Free-Viewpoints Indoor~\cite{philip2021free} and Zip-NeRF~\cite{barron2023zip} have indeed broadened the scope of activity. However, due to the undersampling of the space of camera poses, insufficient density in photo sequences, and the camera lens being fixed in a specific direction (such as always facing forward), achieving 6DoF motion capture becomes a challenge. Consequently, high-quality viewpoint rendering can only effectively be achieved in the vicinity of the data collection trajectory. To illustrate, if you explore a VR scene built from such datasets and deviate from the original data collection trajectory, even by a slight bit, the synthesized novel view you encounter might be of poor quality.

There are also examples of dynamic scene acquisition using monocular cameras, such as ~\cite{yoon2020novel}  , but it with a handheld camera is only capable of recording short-term dynamic events (around 5 seconds) at a resolution of 1920x1080. In summary, regardless of the data acquisition methodology, datasets based on monocular cameras suffer from a lower density of captured viewpoints and either cannot capture dynamic scenes or have limited acquisition time and information about dynamic scenes.

\subsection{Multi-Camera-Based Dataset} Observing multiple perspectives simultaneously allows for the acquisition of additional spatial information and visual context, thereby enhancing the understanding of the behavior of scenes and objects. Consequently, in early studies, the development of multi-camera capture systems was predominantly utilized for capturing human-centric or object-centric activities in indoor environments. The Human3.6M dataset~\cite{ionescu2013human3}, with its extensive collection of high-resolution images, became a benchmark for human pose estimation. The CMU Panoptic Studio~\cite{joo2015panoptic}, a trailblazer in this field, utilized a vast array of cameras to intricately capture social interactions, setting a high standard for detailed motion analysis. Similarly, the MPII Cooking 2 Dataset~\cite{rohrbach2016recognizing} expanded the scope by capturing a wider range of human activities. In recent years, datasets like ~\cite{mahmood2019amass,punnakkal2021babel,guo2022generating,lin2023motion,voynov2023multi} have also leveraged the advantages of multi-camera systems to enhance the richness of data. They are now widely used in fields such as human pose estimation, the generation of 3D human models and 3D object reconstruction.

With technological advancements, multi-camera setups expanded their scope to include entire scenes, leading to a new wave of datasets focused on 3D scene reconstruction. SUN3D~\cite{xiao2013sun3d} provided a wealth of indoor scenes, aiding significantly in indoor space understanding. Matterport3D~\cite{chang2017matterport3d}, with its detailed RGB-D panoramas, became invaluable for indoor 3D modeling. Yet, both datasets were limited by their focus on static scenes, not accounting for dynamic changes. Datasets like ETH3D~\cite{schoeps2017cvpr} and DeepView~\cite{flynnDeepViewViewSynthesis2019} further diversified the scope by covering both indoor and outdoor scenes. ETH3D used a DSLR camera as well as a synchronized multi-camera rig with varying field-of-view to capture images, being beneficial for building 3D reconstruction benchmark. DeepView focused on view synthesis, capturing 100 indoor and outdoor scenes with a 16-camera rig at a 2K resolution. But they were limited by having fewer images, and primarily covered simple, limited-space scenarios. The Immersive Light Field dataset~\cite{broxtonImmersiveLightField2020b}, represents a significant advancement on previous studies, specializing in immersive light field videos captured through a hemispherical array of 46 cameras. Nevertheless, it remains restricted to fixed-point capture, failing to meet the demands for a free and in-depth exploration of reconstructed virtual scenes.

The most recent progression in this field is epitomized by datasets like VR-NeRF~\cite{xuVRNeRFHighFidelityVirtualized2023}, reflecting a trend towards capturing larger, more complex scenes, particularly for virtual reality applications. Utilizing 21-22 cameras with high-resolution imagery, VR-NeRF stands out for its full coverage of large indoor spaces, making it highly suitable for VR environments and complex scene reconstructions. However, its focus on indoor scenes and a less dense data capture methodology limit its application for outdoor environments or scenarios requiring more intensive data capture to obtain higher precision and greater details.

To address these issues and meet the growing demand for 3D scene data, we designed the capture system and capture strategy that integrates the strengths of the above studies. We built the datasets named Den-SOFT focusing on high-resolution, high-density capturing of the large-scale indoor and outdoor scenes.

\section{Den-SOFT Dataset}

\begin{figure*}
    \centering
    \begin{subfigure}{0.3\textwidth}
        \includegraphics[width=\textwidth]{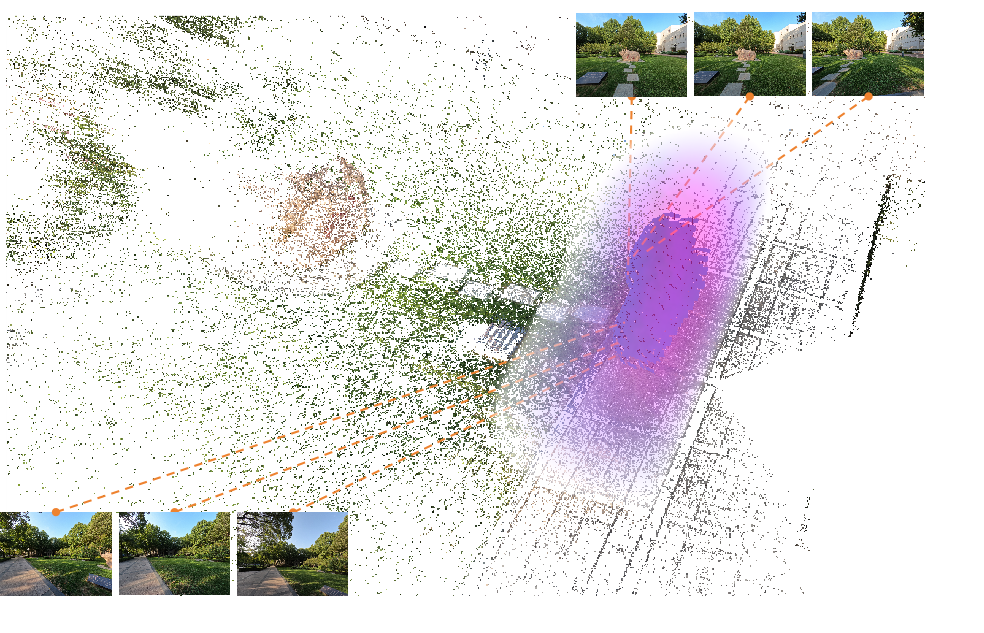}
        \caption{RuziNiu Statue}
    \end{subfigure}
    \hfill
    \begin{subfigure}{0.3\textwidth}
        \includegraphics[width=\textwidth]{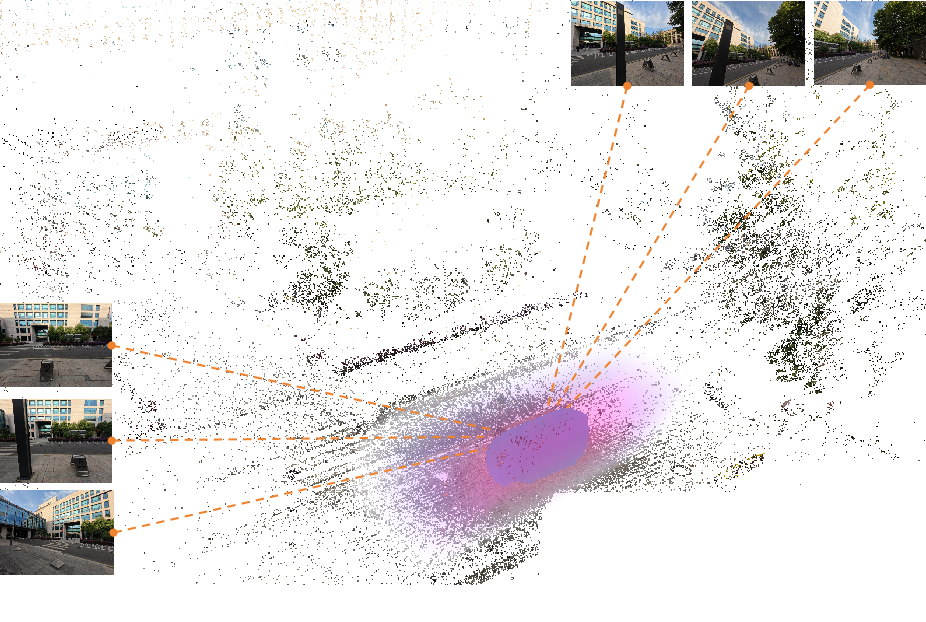}
        \caption{LizhaoJi Building}
    \end{subfigure}
    \hfill
    \begin{subfigure}{0.3\textwidth}
        \includegraphics[width=\textwidth]{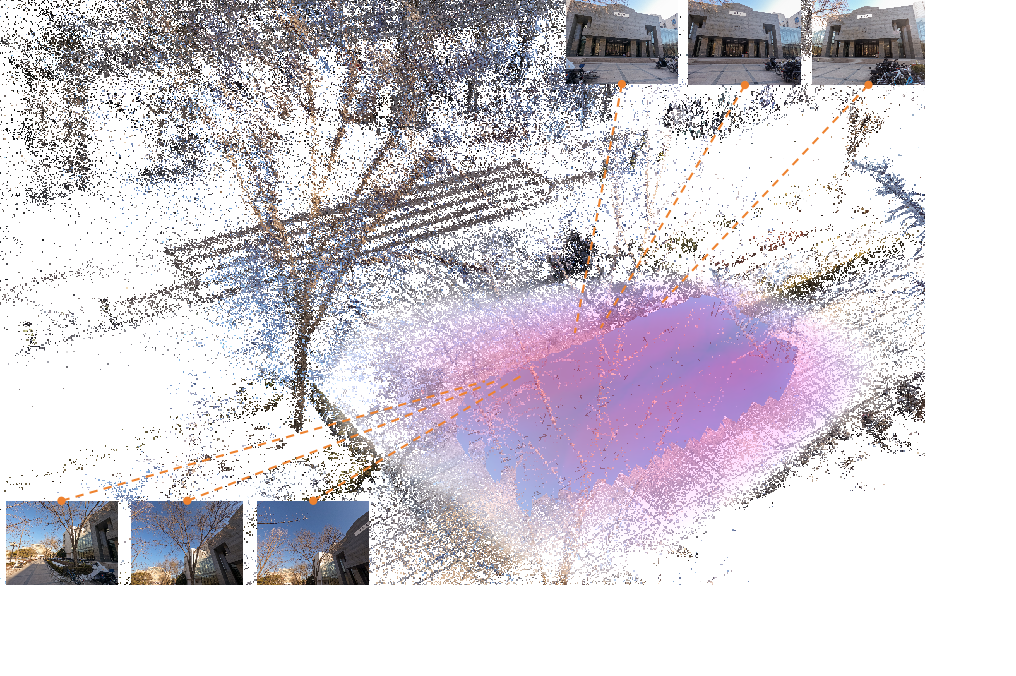}
        \caption{Architecture}
    \end{subfigure}
    
    \begin{subfigure}{0.3\textwidth}
        \includegraphics[width=\textwidth]{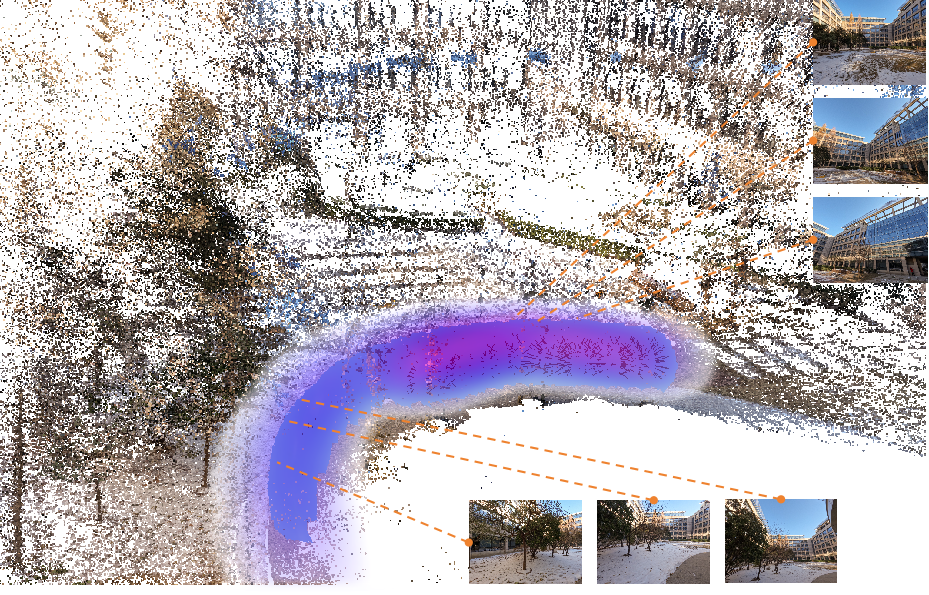}
        \caption{Center Garden}
    \end{subfigure}
    \hfill
    \begin{subfigure}{0.3\textwidth}
        \includegraphics[width=\textwidth]{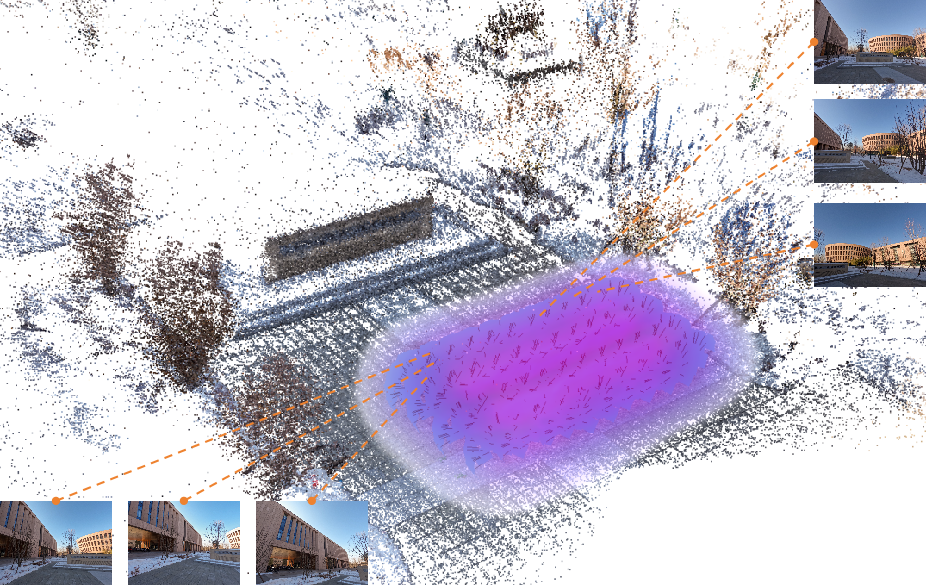}
        \caption{Square1}
    \end{subfigure}
    \hfill
    \begin{subfigure}{0.3\textwidth}
        \includegraphics[width=\textwidth]{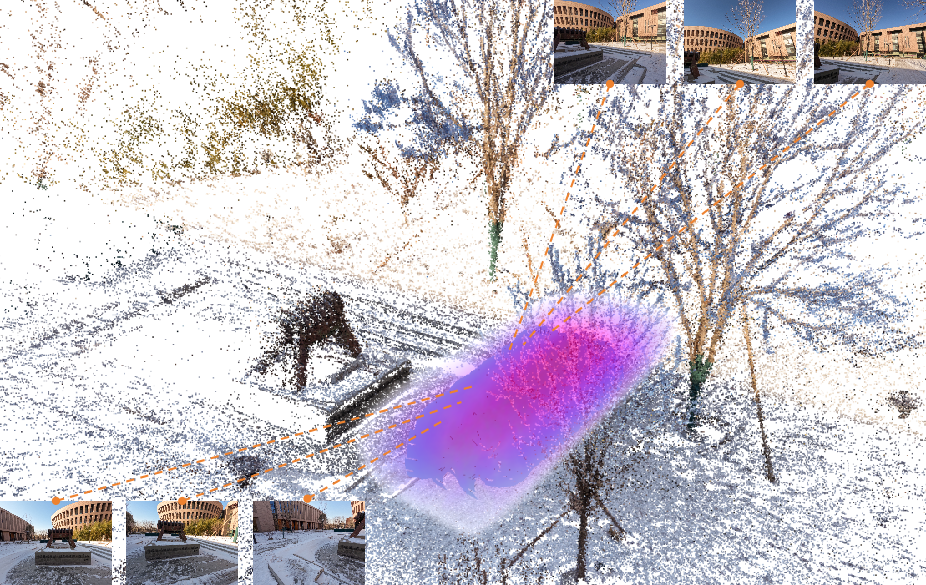}
        \caption{Square2}
    \end{subfigure}

    \begin{subfigure}{0.3\textwidth}
        \includegraphics[width=\textwidth]{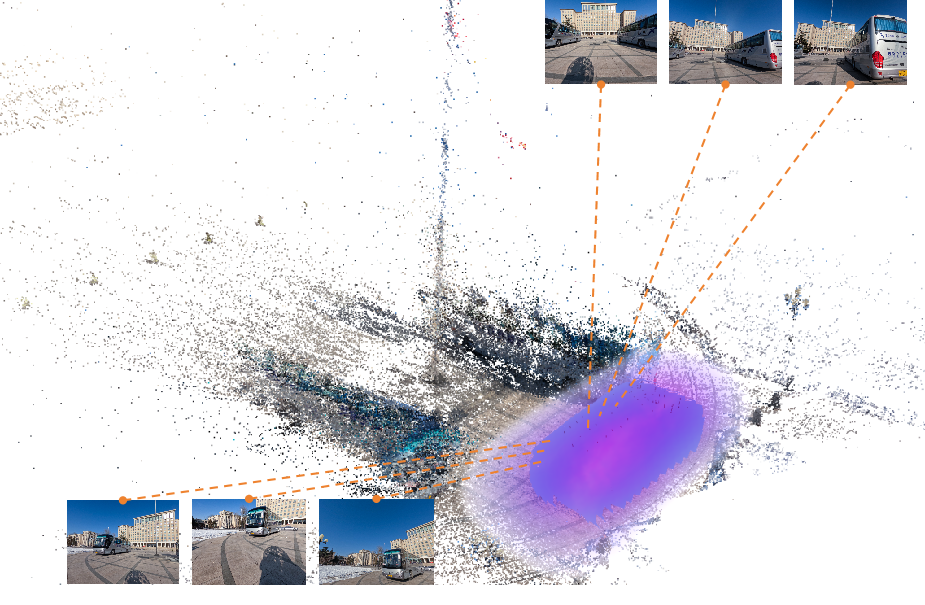}
        \caption{Flagstaff}
    \end{subfigure}
    \hfill
    \begin{subfigure}{0.3\textwidth}
        \includegraphics[width=\textwidth]{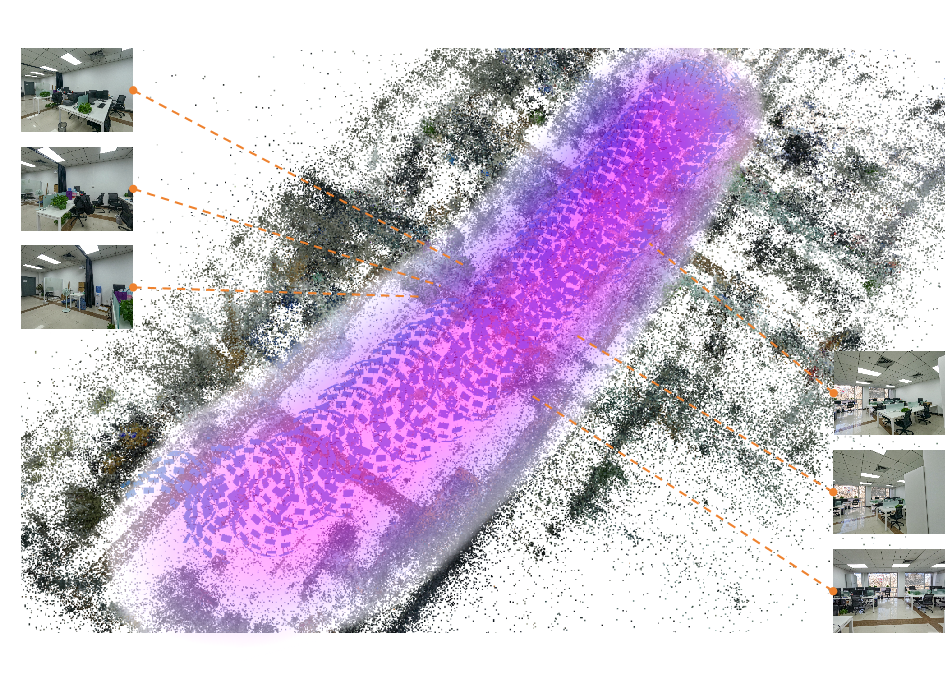}
        \caption{Office1}
    \end{subfigure}
    \hfill
    \begin{subfigure}{0.3\textwidth}
        \includegraphics[width=\textwidth]{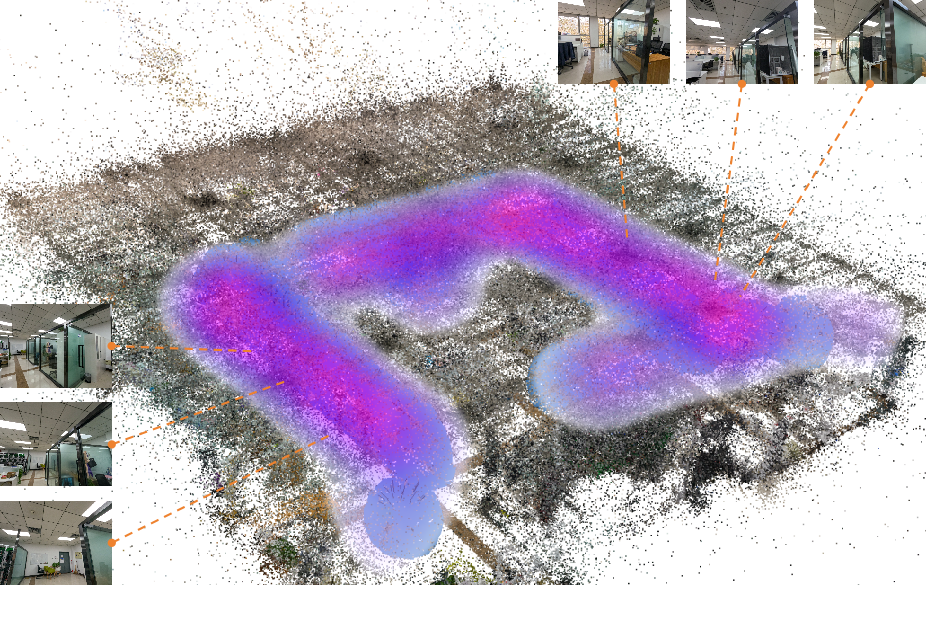}
        \caption{Office2}
    \end{subfigure}
    \caption{A glance of raw data, camera poses and capture density of our Den-SOFT datasets. We collected data from a total of nine scenarios and visualized the capture volume. The sampling density from blue to red indicates sparse to dense.}
    \label{fig:dataset overview}
\end{figure*}

The goal of Den-SOFT dataset is to provide a high-quality and challenging benchmark to the field of large-scale scene reconstruction. Sec 3.1 describes the design of capture rig. Sec 3.2 and Sec 3.3 provide detailed procedure of capturing and data processing. Sec 3.4 presents the statistics and characteristics of this dataset.

\subsection{Mobile Light Field Capturing Device}

\begin{wrapfigure}{r}{0.4\columnwidth}
  \centering
  \includegraphics[width=0.4\columnwidth]{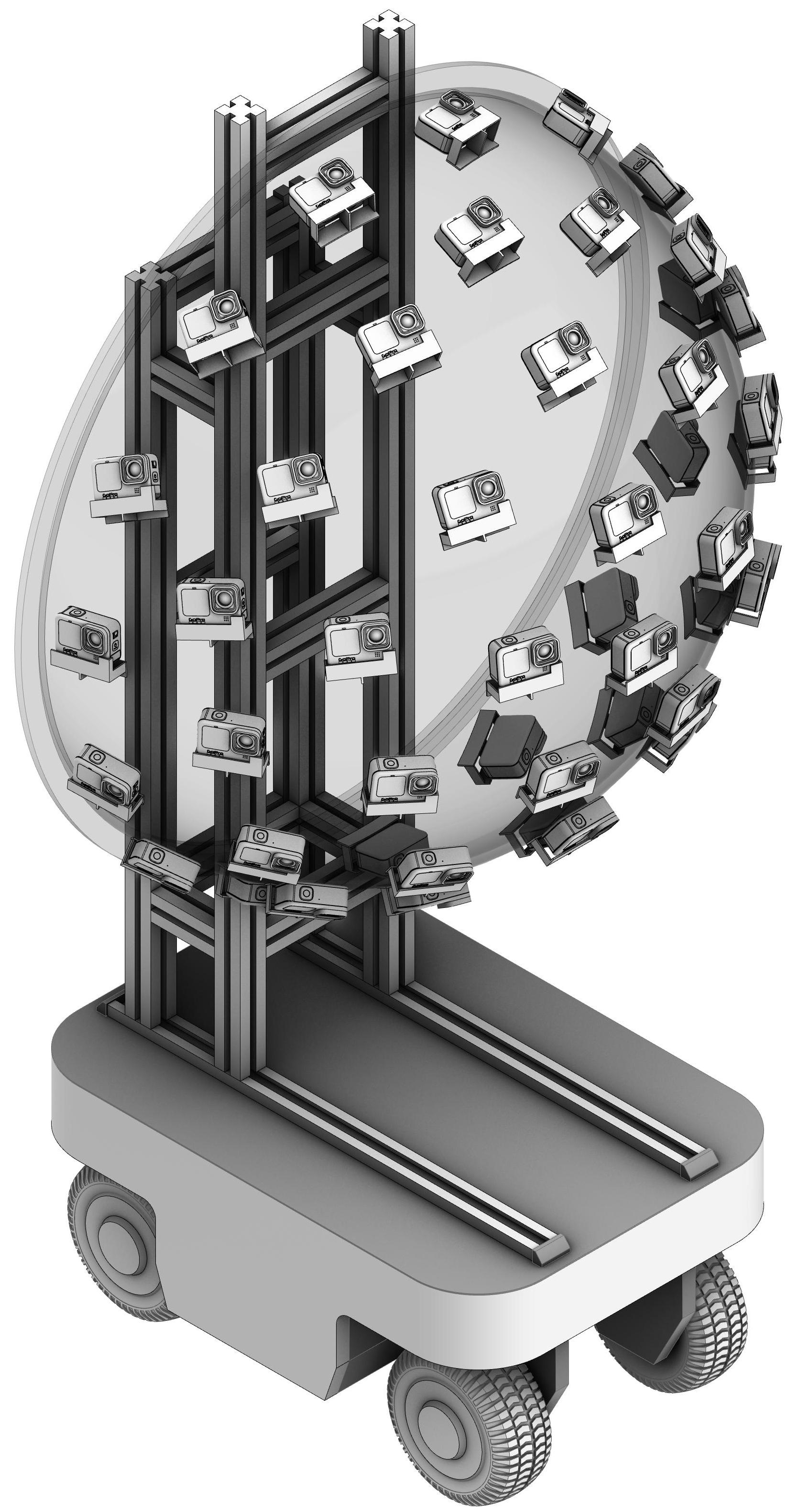}
\end{wrapfigure}

In order to efficiently capture light field with high fidelity, we have come up with a customized light field capturing device as shown in the figure. The design resembles a compound eye with a maximum amount of 46 cameras evenly distributed across the surface. The design is driven by the initial ambition that one can rapidly scan through large scale scenes by 'walking' through it, without sacrificing details and key perspectives that eventually determines the level of immersion. 

The system therefore employs a very robust and straight-forward logic to construct: 46 small yet powerful sports camera spread across a 9200mm-diameter transparent acrylic dome, which allows easy observation of every camera screen from a controller perspective. Dome itself then sits firmly on a mobile chassis, having the ability to carry a maximum load of 80kg while moving freely through sometimes challenging terrains. The entire system is powered by one large portable battery, supporting long shooting period of times. 

We initially used Fisheye mode for shooting and found that although the perspective range was larger than other modes, there were certain challenges for various software to handle distortion in such data. In order to minimize the impact of distortion on scene reconstruction, we choose to use linear camera mode for acquisition.

We have set up all cameras on the capture rig to ensure consistent parameters, and achieved synchronous acquisition by writing code and controlling the capturing with a computer.

\subsection{Den-SOFT Dataset Capture}
Although more and more people are joining the work of building 3D scene reconstruction datasets with the iteration and upgrade of collection equipment, there is still no good metric to measure whether the collection method is reasonable and effective, especially for large-scale scene reconstructions. It is unknown how much denser photography is needed to achieve better results. Therefore, before actually conducting the scene photography, we first conduct tests on a unreal dataset\cite{broxtonImmersiveLightField2020b}. By observing the effects of different sampling methods on the 3D reconstruction results, we finally decide to use the number of viewpoints within a sphere with r=1 m as a metric to evaluate whether the current collection is sufficiently dense.

We achieve the collection of indoor and outdoor large scenes by controlling the mobile cart. To ensure sufficient collection density, we maintained a collection every 30cm based on the characteristics of different scenes, using the tile lines or markings on the floor as guides. Notably, to fully capture certain scenes, we also implemented 360° photography for specific scenes, encompassing both forward and backward perspectives. During the capture of static scenes, we made every effort to prevent staff from appearing in any camera's line of sight and avoided large-scale movement of objects within the scene. To enhance the richness of the data, we also focused on capturing details like changes in light and shadow, and glass reflections during the collection process. The schematic diagram of the dataset we constructed is shown in Figure~\ref{fig:dataset overview}.

\subsection{Data Processing}
An automatic process is employed to obtain additional information, including camera intrinsic and extrinsic, depth map, and point clouds. 

Due to GoPro cameras automatically focusing in response to changes in lighting during movement, the collected data sometimes exhibits inconsistent brightness and darkness in some areas. Therefore, we use Adobe LightRoom to adjust the overall white balance of the images. For camera poses and intrinsic parameters estimation, since our data volume is large and we seek more precise calibration, we use Agisoft Metashape Pro 2.0.2, a professional photogrammetry software, for calibration. This software can automatically recognize the camera model being used, thereby adding intrinsic priors, significantly reducing computation time and enhancing reconstruction accuracy.

\subsection{Captured Datasets} 
Table~\ref{tab:dataset overview}  provides a detailed presentation of the data collection results for nine scenes using our self-built acquisition device “Compound Eye”. Among them, there are 7 outdoor large-scale scene data and 2 indoor large-scale scene data. We also calculate the size of the capture volume and provide the number of viewpoints per cubic meter, as shown in the second to last and third to last columns of the table. Our capturing density ranges from 88 to 189 viewpoints within a unit sphere.

In order to provide a clearer and more intuitive view of the characteristics of our dataset, we also compare it with commonly used datasets in the field of 3D scene reconstruction. The comparison is shown in Table~\ref{tab:dataset comparison}. Our dataset has achieved a resolution of 5K for all scenes and is capable of capturing photos and videos (i.e. capturing static and dynamic scenes). The dataset includes both large-scale indoor and outdoor scenes and the density of our data collection in unit space is also higher than that of several commonly used datasets currently available.

\begin{table}
    \caption{More details of our datasets. \textbf{VP/m³} column indicates the average number of view points within a unit sphere, which we calculate by randomly sampling 64 × 64 × 64 points in the collection volume. \textbf{Volume} column indicates the capture space size of which at least 10 cameras within 1 m³.}
    \label{tab:dataset overview}
    \resizebox{\columnwidth}{!}{%
    \begin{tabular}{ccccccc}
    \toprule
    Scene             & Camera & Pos & Img   & Volume(m³) & VP/m³  & Size \\ \midrule
    RuziNiu Statue    & 41     & 38  & 1,558 & 11.52      & 135.2  & 19 GB \\
    LizhaoJi Building & 41     & 21  & 861   & 6.83       & 126.1  & 9 GB \\
    Architecture      & 40     & 32  & 1,295 & 12.57      & 103.0  & 13 GB \\
    Center Garden     & 40     & 53  & 2,109 & 23.95      & 88.1   & 23 GB \\
    Square1           & 40     & 40  & 1,588 & 8.37       & 189.7  & 15 GB \\
    Square2           & 40     & 47  & 1,876 & 11.59      & 161.8  & 18 GB \\
    Flagstaff         & 40     & 38  & 1,518 & 8.04       & 188.8  & 15 GB \\
    Office1           & 40     & 43  & 1,710 & 13.14      & 130.1  & 15 GB \\
    Office2           & 40     & 89  & 3,550 & 39.72      & 89.4   & 30 GB \\ \bottomrule
\end{tabular}%
}
\end{table}

\begin{table*}
    \centering
    \caption{Comparison of statistics and properties between our \textbf{Den-SOFT} dataset with previous datasets. \textbf{ETH3D} use 1 DLSR camera for images capture, use 4 global-shutter cameras for videos capture. Its image resolution is 6K, while the video resolution is 752×480. 
    \textbf{Nerf-LLFF} is an object-centric datasets thus we did not classify it into either indoor or outdoor.
    \textbf{DeepView} and \textbf{Immersive Light Field} are both fixed-point capture.
    Avg.Viewpoints indicates the average number of camera poses for a single scene.}
    \label{tab:dataset comparison}
    \resizebox{\textwidth}{!}{%
    \begin{tabular}{cccccccccc}
    \toprule
    Dataset                                                        & Synchronized cameras & Resolution  & Source        & Images       & Videos       & Indoor     & Outdoor      & Avg.Viewpoints        & VP/m³           \\ \midrule
    Nerf-LLFF~\cite{mildenhall2021nerf}                            & 1                    & 4K          & Real          & \Checkmark   & \XSolidBrush & —          & —            & \textless50         & 38.13           \\
    Matterport3D~\cite{chang2017matterport3d}                      & 6                    & 1K          & Real          & \Checkmark   & \XSolidBrush & \Checkmark & \XSolidBrush & \textgreater2k      & 18              \\
    ETH3D~\cite{schoeps2017cvpr}                                   & 1 or 4               & 6K/0.7K     & Real          & \Checkmark   & \Checkmark   & \Checkmark & \Checkmark   & \textless100        & 4.64            \\
    DeepView~\cite{flynnDeepViewViewSynthesis2019}                 & 16                   & 2K          & Real          & \Checkmark   & \XSolidBrush & \Checkmark & \Checkmark   & \textless200        & 140             \\
    Immersive Light Field~\cite{broxtonImmersiveLightField2020b}   & 46                   & 2K          & Real          & \XSolidBrush & \Checkmark   & \Checkmark & \Checkmark   & 46                    & —               \\
    VR-Nerf~\cite{xuVRNeRFHighFidelityVirtualized2023}             & 21 or 22             & 8K          & Real          & \Checkmark   & \Checkmark   & \Checkmark & \XSolidBrush & \textgreater2k      & 104.49          \\
    \textbf{Den-SOFT(ours)}                                        & \textbf{40 or 41}    & \textbf{5K} & \textbf{Real} & \textbf{\Checkmark}                  & \textbf{\Checkmark}                & \textbf{\Checkmark}                & \textbf{\Checkmark}                  & \textbf{\textgreater1.5k} & \textbf{134.68} \\ \bottomrule
\end{tabular}%
}
\end{table*}

\section{Evaluation on various paradigms of light field reconstruction method} 

To validate the effectiveness and rationality of our dataset, we conduct evaluation by employing currently popular light field reconstruction algorithms. It is worth mentioning that, although there is a wide variety of light field reconstruction algorithms available, they can generally be categorized into three paradigms: image-based rendering, neural radiance field based reconstruction, and explict neural radiance filed (with point cloud) based reconstruction. 

We evaluate the rendering performance of three representative method based on PSNR, SSIM and LPIPS, using Nvidia A100 GPU and RTX3090 GPU devices.

\begin{figure*}
    \centering
    \includegraphics[width=\textwidth]{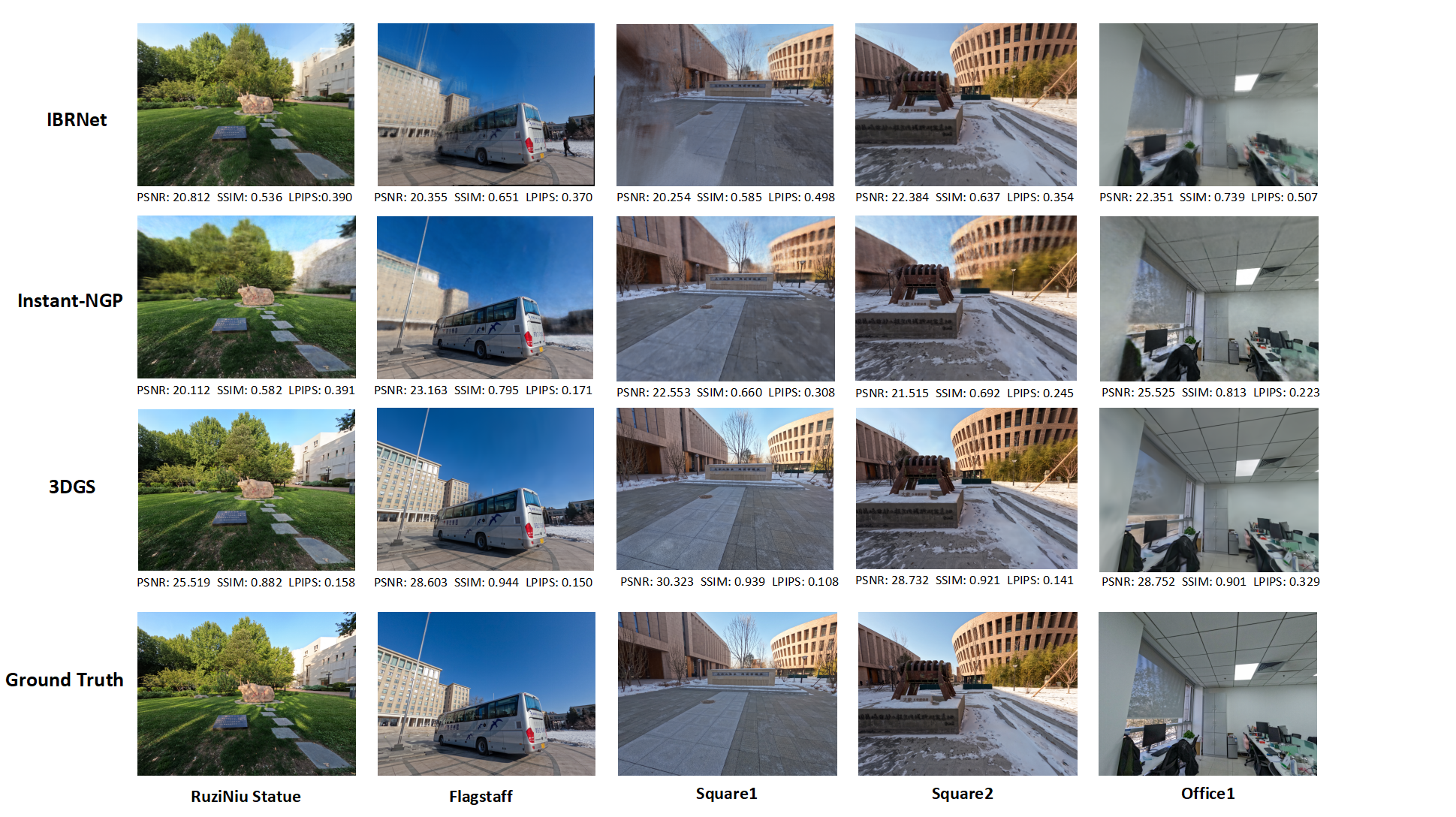}
    \caption{Visualization of Light field reconstruction results of three paradigms method: IBRNet, Instant-NGP and 3D Gaussian Splatting.}
    \label{fig:algorithms comparison}
\end{figure*}

\begin{table}[]
\caption{ Performance comparison of three paradigms algorithms on our datasets.}
\label{tab:tested via various algorithms}
\resizebox{\columnwidth}{!}{%
\begin{tabular}{cccccccccc}
\toprule
\textbf{Scene}       & \multicolumn{3}{c}{\textbf{IBRnet}}                                & \multicolumn{3}{c}{\textbf{Instant-NGP}}                           & \multicolumn{3}{c}{\textbf{3DGS}}                                  \\ \midrule
                     & PSNR↑                & SSIM↑                & LPIPS↓               & PSNR↑                & SSIM↑                & LPIPS↓               & PSNR↑                & SSIM↑                & LPIPS↓               \\
RuziNiu Status       & 18.59                & 0.490                & 0.436                & 19.86                & 0.504                & 0.445                & 24.41                & 0.858                & 0.182                \\
Flagstaff            & 16.49                & 0.507                & 0.572                & 21.68                & 0.724                & 0.233                & 26.54                & 0.903                & 0.177                \\
Square1              & 16.56                & 0.556                & 0.526                & 21.35                & 0.681                & 0.282                & 27.28                & 0.912                & 0.176                \\
Square2              & 17.70                & 0.577                & 0.461                & 22.08                & 0.712                & 0.238                & 27.33                & 0.927                & 0.150                \\
Office1              & 19.61                & 0.719                & 0.552                & 22.14                & 0.709                & 0.365                & 26.88                & 0.883                & 0.296                \\ \bottomrule
\multicolumn{1}{l}{} & \multicolumn{1}{l}{} & \multicolumn{1}{l}{} & \multicolumn{1}{l}{} & \multicolumn{1}{l}{} & \multicolumn{1}{l}{} & \multicolumn{1}{l}{} & \multicolumn{1}{l}{} & \multicolumn{1}{l}{} & \multicolumn{1}{l}{} \\
\multicolumn{1}{l}{} & \multicolumn{1}{l}{} & \multicolumn{1}{l}{} & \multicolumn{1}{l}{} & \multicolumn{1}{l}{} & \multicolumn{1}{l}{} & \multicolumn{1}{l}{} & \multicolumn{1}{l}{} & \multicolumn{1}{l}{} & \multicolumn{1}{l}{} \\
\multicolumn{1}{l}{} & \multicolumn{1}{l}{} & \multicolumn{1}{l}{} & \multicolumn{1}{l}{} & \multicolumn{1}{l}{} & \multicolumn{1}{l}{} & \multicolumn{1}{l}{} & \multicolumn{1}{l}{} & \multicolumn{1}{l}{} & \multicolumn{1}{l}{} \\
\multicolumn{1}{l}{} & \multicolumn{1}{l}{} & \multicolumn{1}{l}{} & \multicolumn{1}{l}{} & \multicolumn{1}{l}{} & \multicolumn{1}{l}{} & \multicolumn{1}{l}{} & \multicolumn{1}{l}{} & \multicolumn{1}{l}{} & \multicolumn{1}{l}{} \\
\multicolumn{1}{l}{} & \multicolumn{1}{l}{} & \multicolumn{1}{l}{} & \multicolumn{1}{l}{} & \multicolumn{1}{l}{} & \multicolumn{1}{l}{} & \multicolumn{1}{l}{} & \multicolumn{1}{l}{} & \multicolumn{1}{l}{} & \multicolumn{1}{l}{}
\end{tabular}%
}
\end{table}

\subsection{Image-Based Rendering}

Image-based rendering (IBR) is a commonly used method when it comes to scene reconstruction. Its main idea is to synthesize novel views by utilizing existing image data, rather than building 3D scenes from scratch. It leverages the information present in images to infer the geometry and appearance properties of a scene. Given that we have collected data from sufficiently dense viewpoints, which theoretically aligns well with the nature of Image-based rendering methods, we initially test our dataset on these types of methods.

IBRNet\cite{wangIBRNetLearningMultiView2021a}, or Image-based Rendering Network, is a cutting-edge approach in the field of Image-based Rendering that integrates deep learning. It uses a deep neural network to synthesize new views from existing images, learning from a vast dataset to accurately infer the geometric and photometric properties of a scene. This method excels in complex scenarios, like occlusions or intricate textures, by predicting how objects should appear from various angles. 

After converting all the scene datas into LLFF format, we test them using the IBRNet pre-trained model and finetune certain scenes as necessary. The training for each scene takes about 6 hours on a single A100, with the results shown in Table~\ref{tab:tested via various algorithms}.

\subsection{Neural Radiance Field Reconstruction}
While the results from IBRnet are reasonable, they did not meet our expectations. We began to seek another paradigm of methods for testing, in the hope of achieving better results. NeRF, short for Neural Radiance Fields\cite{mildenhall2021nerf}, is a novel deep learning framework used for creating photorealistic 3D models from 2D images. It works by mapping 3D coordinates and viewing angles to color and density, using a neural network. This technique allows it to capture complex lighting and realistic details, making it ideal for applications in virtual reality, film, and heritage preservation.

Instant Neural Graphics Primitives (Instant-NGP)\cite{mullerInstantNeuralGraphics2022d}, improves upon the original NeRF by offering much faster training and rendering. Key advancements include a multi-resolution hash table for efficient spatial encoding, optimized neural network architecture for speed, and better scalability for complex scenes. These improvements make Instant-NGP ideal for real-time applications in VR, AR, and gaming. Therefore, we chose it as the second representative method for our experiments.

It should be noted that, although Instant-NGP has greatly improved in terms of training and rendering speed, it still requires significant computational resources when dealing with a large number of images. Therefore, we alter our evaluation strategy, reducing the data while maintaining the same input viewpoint density for testing. Each scene has about 230 images, and on an RTX 3090 graphics card, it takes about 3 minutes on average for 35,000 iterations. The results are also displayed in Table~\ref{tab:tested via various algorithms}.

\subsection{3D Gaussian Splatting Reconstruction}

3D Gaussian Splatting (3DGS)\cite{kerbl3DGaussianSplatting2023b} is the latest method that combines images data, point clouds data, and camera poses to achieve high-quality real-time ($\geq$ 30 fps) novel view synthesis at 1080p resolution. It introduces a three-dimensional Gaussian distribution for scene representation. Starting from the initial sparse point clouds, the method optimizes various properties of 3D Gaussian (such as position, opacity, anisotropic covariance, and spherical harmonic coefficient) and selects to add or remove 3D Gaussian based on density to obtain a compact, unstructured, and accurate scene representation. The process culminates in fast rasterization for image rendering.

We use NVIDIA A100 for training and the training time for each scenario ranges from 40 to 60 minutes. It's important to highlight that the original algorithm does not support the reconstruction of unbounded large scenes,especially in the case of sky reconstruction, where a "collapse" effect may occur. To address this, we calculate the range of the scene and artificially add spherical boundaries to the initial point cloud. These points are then adjusted within the algorithm to ensure they participate in the optimization during training instead of being pruned. This step has proven to be crucial and effective.

\subsection{Result and Analysis}
In conclusion, all three scene reconstruction methods discussed here—-IBRNet, Instant-NGP, and 3D Gaussian Splatting, heavily rely on the quality of input images and the accuracy of camera calibration, especially IBRNet. As indicated by the test results in Table~\ref{tab:tested via various algorithms} and Figure~\ref{fig:algorithms comparison}, our dataset essentially reflects the current limits of using these methods for large-scale 3D scene reconstruction. Firstly, due to the dense nature of our captured data, IBRNet can achieve comparable results to Instant-NGP from certain viewpoints. However, its drawback is also evident, as it struggles with significant color differences between adjacent images due to varying focus during capture, leading to noticeable "block" in the output. Secondly, while Instant-NGP generally outperforms IBRNet in most scenarios, its rendering results show better reconstruction in the foreground but blurred backgrounds. In contrast, IBRNet retains clarity in distant views, and in some cases, its rendering quality for high-reflectance objects like glass even surpasses that of 3D Gaussian Splatting (3DGS). Lastly, both IBRNet and Instant-NGP are inferior to 3DGS in our dataset's performance. 3DGS might be the most suitable method currently for large-scale scene reconstruction, but we also observe its limitations when dealing with strong lighting and highly reflective objects.

\begin{figure}
    \centering
    \includegraphics[width=\columnwidth]{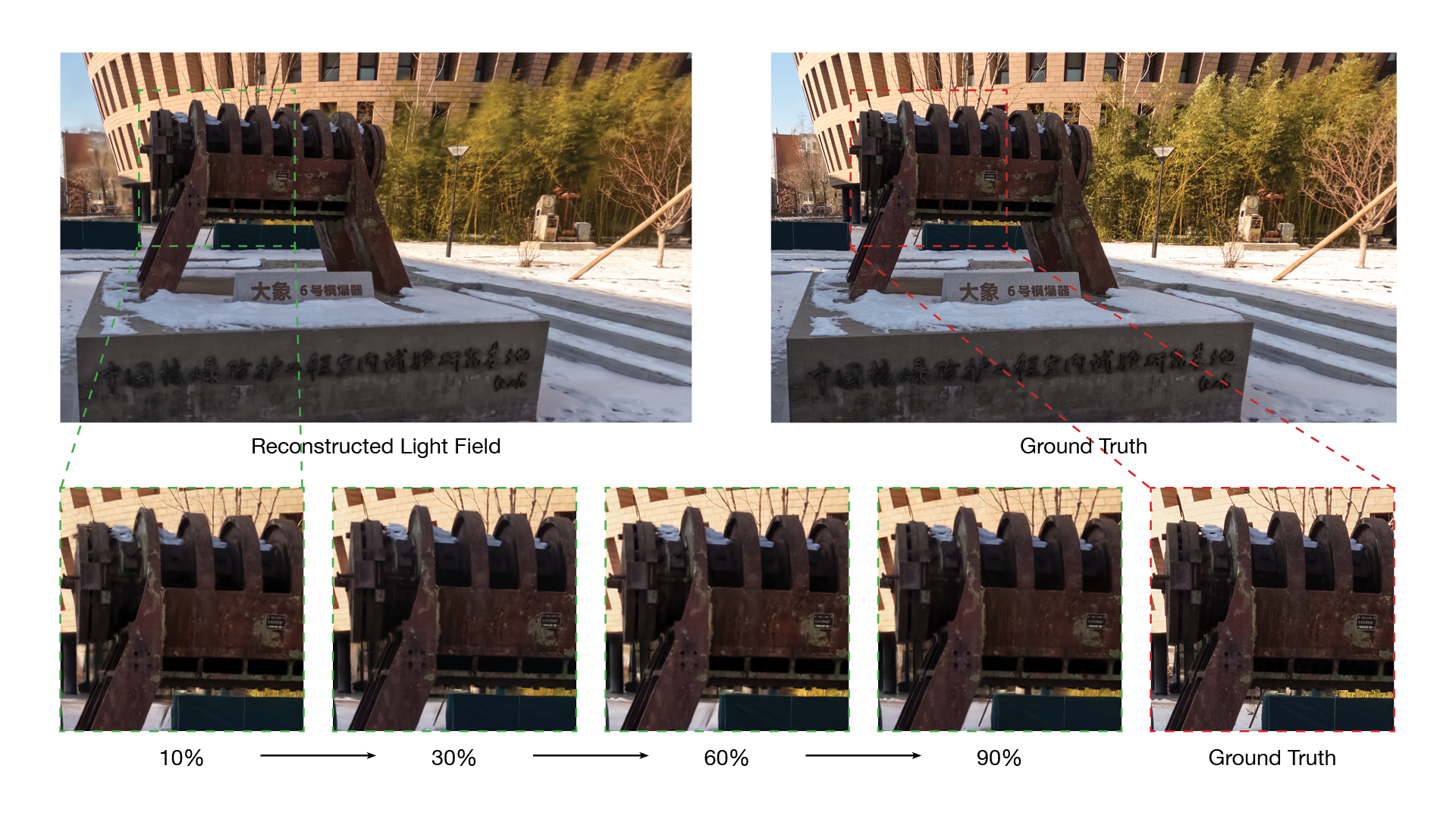}
    \caption{3D Gaussian Splatting reconstruction results under different image density input.}
    \label{fig:Density comparison}
\end{figure}

\begin{figure}
  \centering
  \includegraphics[width=\columnwidth]{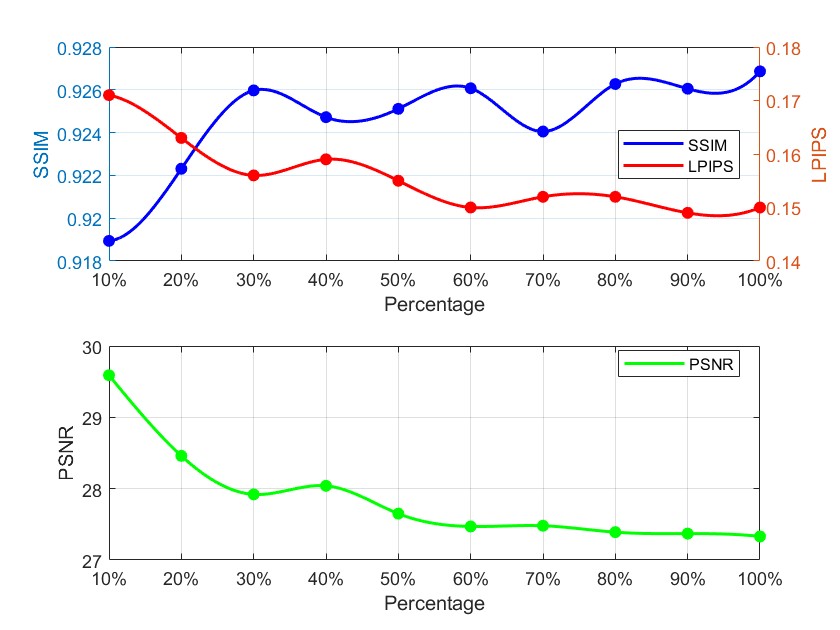}
  \caption{3DGS performance under different capture density.}
  \label{fig:multi}
\end{figure}

In addition, due to the large number of images we have collected and the high density of our data acquisition, our dataset may be able to assess the most suitable capture density of large-scale scenes for 3D reconstruction algorithms. We conducted tests on the 3DGS algorithm for the \textit{Square2} scene, and the results are shown in Table~\ref{tab:test on different density} and Fig ~\ref{fig:Density comparison}. From the preliminary results and the curve analysis shown in Fig~\ref{fig:multi}, it can be seen that as the input data (i.e. sampling density) increases, there is an improvement in the details of the reconstructed scene, but more is not necessarily better. There may be a certain threshold that is worth further exploration.

\begin{table}[]
\caption{Test algorithm on different input image number which represent the various capture density. }
\label{tab:test on different density}
\resizebox{\columnwidth}{!}{%
\begin{tabular}{cccccc}
\toprule
Scene:Square2 & \multicolumn{3}{c}{Method:3DGS} & \multirow{2}{*}{points\_number} & \multirow{2}{*}{images\_number} \\
Density       & PSNR↑    & SSIM↑     & LPIPS↓   &                                 &                              \\ \midrule
100\%         & 27.32    & 0.92686   & 0.150    & 821,793                         & 1,874                        \\
90\%          & 27.37    & 0.92605   & 0.149    & 754,775                         & 1,688                        \\
80\%          & 27.39    & 0.92627   & 0.152    & 695,546                         & 1,499                        \\
70\%          & 27.48    & 0.92405   & 0.152    & 607,813                         & 1,310                        \\
60\%          & 27.47    & 0.92607   & 0.150    & 553,538                         & 1,120                        \\
50\%          & 27.65    & 0.92511   & 0.155    & 483,691                         & 937                          \\
40\%          & 28.04    & 0.92472   & 0.159    & 392,567                         & 750                          \\
30\%          & 27.92    & 0.92597   & 0.156    & 315,031                         & 562                          \\
20\%          & 28.46    & 0.92231   & 0.163    & 228,458                         & 375                          \\
10\%          & 29.59    & 0.91894   & 0.171    & 129,499                         & 182                          \\ \bottomrule
\end{tabular}%
}
\end{table}

\section{Dataset Application and Extension}
 By capturing enough density multi-view images of large-scale scenes, our dataset facilitates a more detailed understanding of the scene geometry, appearance, and spatial relationships. In this section, we explore the various applications and potential extensions of our dataset. 
 
\begin{figure}
    \centering
    \includegraphics[width=\columnwidth]{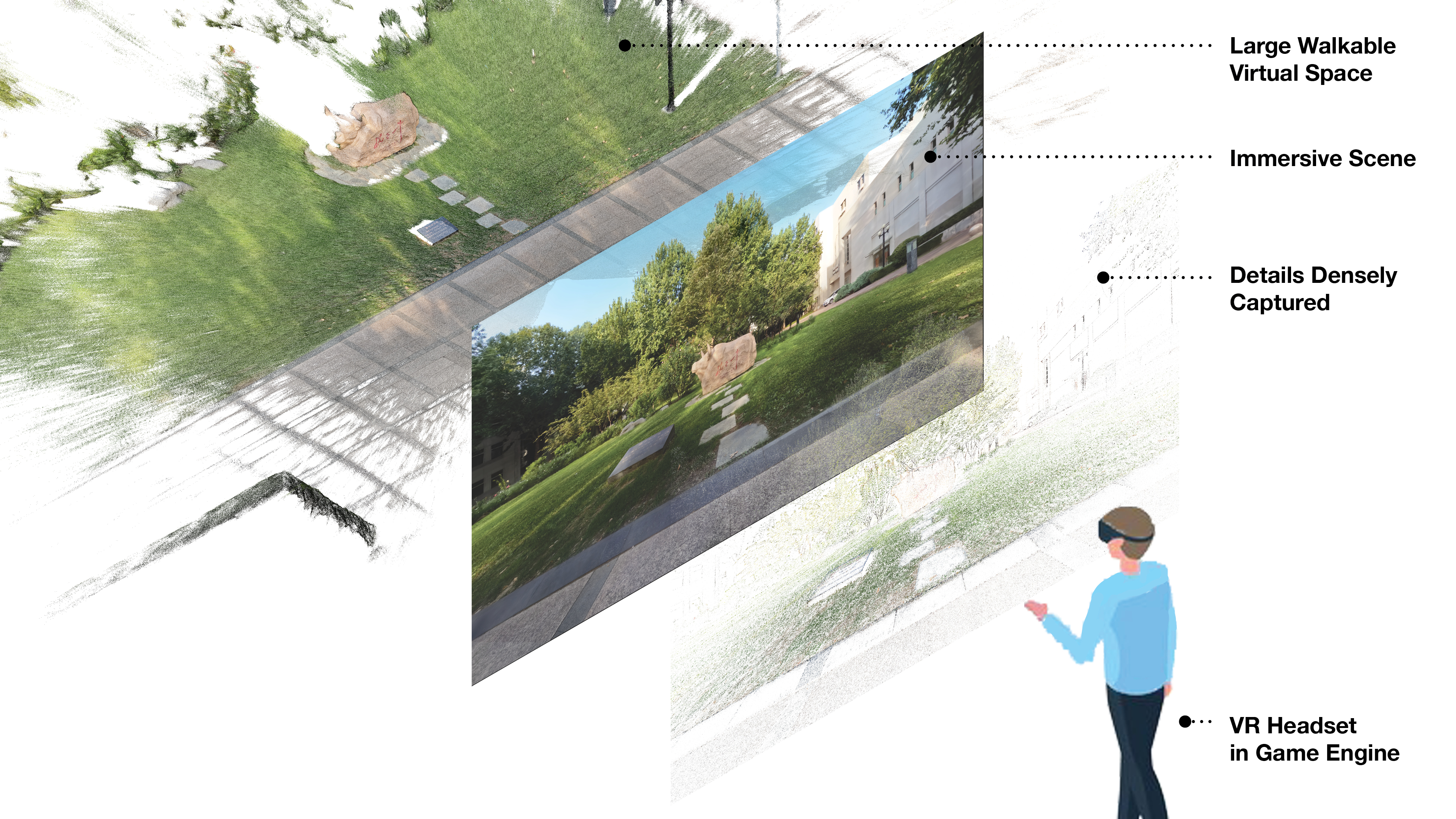}
    \caption{Den-SOFT's potential application: users can import training results into game engines for a six-degrees-of-freedom immersive VR experience.}
    \label{fig:VR-Application}
\end{figure}

\textbf{Immersive and Interactive VR Scenes.} Our dataset plays a crucial role in enabling immersive and interactive virtual reality (VR) experiences. By capturing multi-view images of large-scale scenes, users can navigate and interact within a realistic virtual environment. The dataset provides diverse viewpoints, enhancing the immersion and engagement of the VR experience. We achieved satisfactory results by integrating the training results of the dataset with 3DGS into the Unity rendering engine, as shown in Figure~\ref{fig:VR-Application}. In the future, researchers can leverage this dataset to develop innovative VR applications, such as virtual tourism, architectural walkthroughs, or training simulations. 

\textbf{Dynamic Scene Reconstruction.} Dynamic scene reconstruction is an evolving field with promising future directions. Our designed capture system has the capability to rapidly and continuously capture sequences of dynamic scenes. We are capable of creating a collection space for specific scenes through shooting in a short period of time, which is other current research cannot achieve. Therefore, a potential area of our future research is the development of robust algorithms that can handle challenging scenarios, such as scenes with fast-moving objects or occlusions, by leveraging self-captured datasets.

\textbf{Contribute to Other Promising Tasks.} With sufficiently dense collection of scene data, our work holds potential to bolster a diverse array of light field tasks in the future. One avenue for exploration is the creation of scalable algorithms that can tackle the challenges posed by the sheer volume of data inherent in large-scale scenes. This includes efficient data representation, feature extraction, and inference techniques that can handle the complexity and diversity of such scenes. Another promising direction is the incorporation of contextual information such as scene semantics, temporal dynamics or spatial relationships, to enhance the machine's understanding of real-world scenes. Moreover, the dataset we provide may further stimulate the development of the communication field. For example, if integrated real-time data streams can be achieved, researchers may realize light field reconstruction with real-time data streams fed from sensor networks or social media which provide current information about the scene, this will greatly promote the implementation and application of real-time dynamic scene reconstruction.

\section{Conclusion}

We present Den-SOFT, a dataset focuses on capturing large-scale scenes, especially outdoor unbounded environments, with the aim of providing assistance to the industry in building virtual reality scenes that are more precise, detailed and realistic. By collecting high-resolution image data at high density, we have achieved the reconstruction of a six degree of freedom exploratory space in virtual reality using the currently popular 3DGS method, and also validated the effectiveness of our data on multiple algorithm paradigms. In addition to other contributions, we have independently designed a unique multi-camera rig and reviewed the current technological pipeline, from the capture process to data processing, light field reconstruction, and final rendering and interaction using Unity. We hope that this work can effectively advance the current research on large-scale scene reconstruction.


\bibliographystyle{ACM-Reference-Format}
\bibliography{Den-SOFT_Bib}

\appendix

\end{document}